\documentstyle[tree-dvips,avm,eaclap]{article}
\def\reentry#1{\fboxsep=0ex\framebox[2ex]{\rule[-.6ex]{0ex}{2.3ex}#1}}

\author{Guido Minnen, Dale Gerdemann, Thilo G\"otz
\thanks{
The  presented   research  was  sponsored  by  Teilprojekt  {\sc   b}4
``Constraints   on   Grammar   for   Efficient   Generation''  of  the
Sonderforschungsbereich 340  ``Sprachtheoretische Grundlagen f\"ur die
Computerlinguistik''  of  the  Deutsche  Forschungsgemeinschaft.   The
authors wish to thank Paul King, Detmar Meurers and Shuly Wintner  for
valuable  comments  and   discussion.  Of  course,   the  authors  are
responsible for all remaining errors.  }\\ Sonderforschungsbereich 340
\\University                of                 T\"ubingen\\Wilhelmstr.
113\\72074-T\"ubingen\\Germany\\E-mail:
minnen@sfs.nphil.uni-tuebingen.de}
\title{Off-line Optimization for Earley-style HPSG Processing}
\begin{document}
\maketitle
\begin{abstract}
A novel  approach to {\sc  hpsg} based natural language processing  is
described  that  uses an  off-line compiler to  automatically  prime a
declarative grammar  for generation or parsing, and inputs  the primed
grammar to an advanced Earley-style processor.  This way we provide an
elegant  solution to  the  problems  with  empty  heads and  efficient
bidirectional processing which is illustrated for the special  case of
{\sc  hpsg} generation.  Extensive  testing  with  a large {\sc  hpsg}
grammar  revealed  some  important  constraints  on  the  form  of the
grammar.
\end{abstract}

\section{Introduction}
Bidirectionality of grammar is  a research  topic in  natural language
processing that is enjoying increasing attention \cite{Strzalkowskia}.
This is mainly due to  the clear  theoretical and practical advantages
of bidirectional grammar use (see,  among  others, Appelt,  1987).  We
address  this  topic  in  describing  a novel  approach to {\sc  hpsg}
\cite{Pollard+Sag}  based  language  processing that  uses an off-line
compiler to  automatically prime a declarative grammar  for generation
or  parsing,  and  hands  the  primed grammar  to  an advanced  Earley
processor.  The developed techniques  are direction independent in the
sense that they can be used  for both generation and parsing with {\sc
hpsg} grammars.  In this  paper, we  focus on  the application  of the
developed  techniques in the  context  of the  comparatively neglected
area of {\sc hpsg} generation.

Shieber  (1988)  gave   the  first  use  of  Earley's  algorithm   for
generation,  but this algorithm  does not use  the  prediction step to
restrict feature  instantiations on the  predicted  phrases, and  thus
lacks goal-directedness.  Though Gerdemann (1991) showed how to modify
the  restriction  function to make top-down  information available for
the  bottom-up  completion   step,  Earley  generation  with  top-down
prediction still has  a problem in that generating  the  subparts of a
construction in the  wrong order  might lead to massive nondeterminacy
or even nontermination.  Gerdemann (1991) partly overcame this problem
by  incorporating a  head-driven  strategy  into  Earley's  algorithm.
However, evaluating the head of a construction  prior to its dependent
subparts still suffers  from  efficiency problems when the  head of  a
construction  is  either   missing,   displaced   or   underspecified.
Furthermore,  Martinovi\'{c}  and  Strzalkowski  (1992)  and  others  have
observed that a simple head-first reordering of  the grammar rules may
still  make   insufficient   restricting   information  available  for
generation unless  the  form of the  grammar is restricted to unary or
binary rules.

Strzalkowski's Essential  Arguments Approach ({\sc eaa}; 1993b)  is  a
top-down approach  to generation and  parsing with logic grammars that
uses  off-line  compilation  to automatically  invert  parser-oriented
logic  grammars.  The inversion process consists of both the automatic
static  reordering of nodes in the  grammar, and  the interchanging of
arguments in rules with recursively defined heads.  It is based on the
notion  of  {\it  essential  arguments},  arguments   which   must  be
instantiated to ensure the efficient  and terminating execution  of  a
node.   Minnen  et   al.\  (1995)  observe  that  the  {\sc   eaa}  is
computationally infeasible, because  it  demands  the investigation of
almost  all  possible  permutations  of  a  grammar.   Moreover,   the
interchanging of arguments  in  recursive  procedures as  proposed  by
Strzalkowski  fails  to guarantee that input  and output grammars  are
semantically equivalent.  The Direct Inversion Approach ({\sc dia}) of
Minnen  et  al.\  (1995)  overcomes  these  problems   by  making  the
reordering process more  goal-directed and developing a  reformulation
technique that allows the successful treatment  of rules which exhibit
head-recursion.  Both  the {\sc eaa} and the {\sc  dia} were presented
as approaches  to  the  inversion  of  parser-oriented  grammars  into
grammars  suitable for generation.  However, both approaches  can just
as well take a declarative grammar specification  as  input to produce
generator  and/or  parser-oriented  grammars  as  in Dymetman  et  al.
(1990).   In  this  paper  we  adopt  the  latter  theoretically  more
interesting perspective.

We developed a compiler for off-line optimization of phrase  structure
rule-based  typed feature  structure  grammars  which  generalizes the
techniques developed  in the context of the {\sc dia}, and we advanced
a typed extension  of the Earley-style generator of Gerdemann  (1991).
Off-line  compilation (section  \ref{off-line})  is  used  to  produce
grammars for  the  Earley-style generator (section \ref{earley}).   We
show that our use of  off-line grammar optimization overcomes problems
with  empty   or  displaced   heads.   The  developed  techniques  are
extensively tested with  a large  {\sc hpsg} grammar  for partial {\sc
vp} topicalization in German \cite{HMN}.  This uncovered some important
constraints on the form of the phrase structure rules (phrase structure rules)
in a grammar imposed by the compiler (section \ref{pvp}).
\section{Advanced Earley Generation}
\label{earley}
As  Shieber (1988) noted, the main shortcoming of Earley generation is
a lack of goal-directedness that results in a  proliferation of edges.
Gerdemann (1991) tackled this shortcoming by modifying the restriction
function to  make  top-down  information available for  the  bottom-up
completion step.  Gerdemann's generator follows a head-driven strategy
in order to avoid  inefficient evaluation orders.   More specifically,
the  head of the right-hand side  of each grammar rule is distinguished, and
distinguished  categories  are scanned or predicted  upon first.   The
resulting  evaluation strategy is  similar to that  of the head-corner
approach \cite{Shieberetal,Gerdemann+Hinrichs}: prediction follows the
main  flow of  semantic information  until a lexical pivot is reached,
and only  then  are  the head-dependent subparts  of  the construction
built  up  in  a  bottom-up  fashion.   This  mixture of top-down  and
bottom-up information  flow  is crucial  since  the  top-down semantic
information  from  the  goal  category  must  be integrated  with  the
bottom-up  subcategorization information from  the lexicon.   A strict
top-down  evaluation strategy suffers  from  what  may be called  {\it
head-recursion},  i.e.  the generation  analog  of  left  recursion in
parsing.   Shieber et  al.   (1990)  show that  a  top-down evaluation
strategy will fail for  rules such as {\sc vp} $\rightarrow$  {\sc vp
x},  irrespective   of   the   order   of  evaluation  of   the   right-hand
side
categories in  the rule.   By combining the off-line optimization
process  with  a mixed bottom-up/top-down evaluation strategy,  we can
refrain from a complete reformulation of the grammar  as, for example,
in Minnen et al.\ (1995).

\subsection{Optimizations}
We further improved a  typed extension of Gerdemann's Earley generator
with  a number of  techniques that reduce the number  of edges created
during generation.  Three optimizations were especially helpful.   The
first  supplies each edge  in  the chart  with  two  indices,  a  {\it
backward  index} pointing to the state in the chart  that the edge  is
predicted from, and a {\it forward index} pointing to the states  that
are  predicted  from  the  edge.   By matching  forward  and  backward
indices, the edges that must be combined for completion can be located
faster. This indexing technique, as illustrated below, improves  upon the
more complex indices in Gerdemann (1991)  and  is  closely related  to
{\sc oldt}-resolution (Tamaki and Sato, 1986).\\
\vspace*{-.5ex}

\noindent
1) $ active(X_0 \rightarrow X_1 \bullet X_2,1,\node{21}{2}) $
\nodepoint{i1}[1em][-6ex]

\noindent
\hspace*{5em}\vdots

\noindent
2) $ active(X_2 \rightarrow \bullet\, Y_1\, Y_2,\node{22}{2},3) $

\noindent
\hspace*{5em}\vdots

\noindent
3) $ active(X_2 \rightarrow Y_1\, \bullet\, Y_2,\node{23}{2},4) $

\noindent
\hspace*{5em}\vdots

\noindent
4) $ passive(X_2 \rightarrow Y_1\, Y_2\, \bullet,\node{24}{2}) $
\nodepoint{i2}[1.2em][12ex]

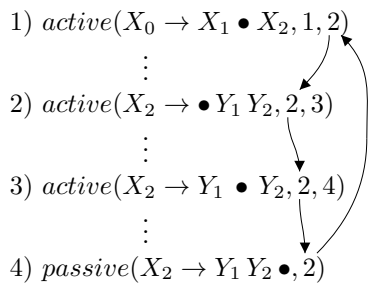
\anodecurve[b]{21}[tr]{22}{2ex}
\anodecurve[b]{22}[t]{23}{2ex}
\anodecurve[b]{23}[t]{24}{2ex}
\anodecurve[t]{i2}[br]{21}{4ex}
\nodecurve[tr]{24}[b]{i2}{4ex}
\vspace*{2ex}

\noindent
Active edge 2 resulted  from active edge  1 through  prediction.   The
backward index of  edge 2 is  therefore  identified  with the  forward
index of edge 1. Completion of  an active edge results in an edge with
identical  backward  index.  In the case of our example, this would be
the steps from edge 2 to edge 3 and edge 3 to edge 4.  As nothing gets
predicted from a passive edge (4),  it does not have a forward  index.
In order to use passive  edge 4  for completion of an  active edge, we
only need to consider those edges which have a forward index identical
to the backward index of 4.

The second optimization  creates a table of  the categories which have
been used to make predictions from.  As discussed in Gerdemann (1991),
such a table can be used to avoid redundant predictions without a full
and expensive subsumption  test.   The third  indexes  lexical entries
which is necessary to obtain constant-time lexical access.

The optimizations of our Earley-generator lead to significant gains in
efficiency.   However,  despite  these   heuristic  improvements,  the
problem of goal-directedness is not solved.

\subsection{Empty Heads}
\label{limitations}
Empty  or  displaced  heads  present  the principal  goal-directedness
problem for any head-driven generation approach (Shieber et al., 1990;
K\"onig, 1994; Gerdemann and  Hinrichs, in  press),  where empty  head
refers not  just  to a  construction in which the  head  has  an empty
phonology,  but to any  construction in  which the head  is  partially
unspecified.    Since  phonology   does  not   guide  generation,  the
phonological realization of  the head of a construction plays no  part
in  the generation  of that construction.  To  better  illustrate  the
problem that underspecified heads pose, consider the sentence:

{\it {\flushleft Hat} Karl Marie gek\"u{\ss}t?}\\
Has Karl Marie kissed?\\
``Did Karl kiss Mary?''

{\flushleft for}  which  we  adopt the  argument composition  analysis
presented  in Hinrichs  and Nakazawa  (1989): the  subcat list of  the
auxiliary verb is  partially  instantiated  in  the  lexicon  and only
becomes  fully  instantiated  upon  its  combination  with  its verbal
complement,  the main verb.   The phrase structure rule that  describes  this
construction is
\footnote{
For expository reasons, we refrain from a division between the subject
and the other complements of a verb as in chapter 9 of Pollard and Sag
(1994).   The  test-grammar  does  make   this  division   and  always
guarantees the correct  order  of the complements  on the  {\it comps}
list with respect to the  obliqueness  hierarchy.  Furthermore, we use
abbreviations    of   paths,   such   as   {\it    cont}   for    {\it
synsem$|$loc$|$cont},  and assume  that  the  semantics  principle  is
encoded in the phrase structure rule.}

\begin{scriptsize}
\begin{center}
\begin{avm}
\[ cat & v\\
   subcat & \<\>\\
   cont & \@5
\]
$\longrightarrow$\end{avm}
\newline\begin{avm}
\[ cat & v\\
   fin & $+$\\
   aux & $+$\\
   subcat & \<\@3\|\@4\>\\
   cont & \@5
\]
,
\@1
,
\@2
,
\@3\[cat & v\\
     fin & $-$\\
     aux & $-$\\
     subcat & \@4\<\@1\|\@2\>
\]
\end{avm}
\end{center}
\end{scriptsize}
Though a  head-driven generator must generate first  the  head of  the
rule, nothing prescribes the order of generation of the complements of
the  head.   If the generator generates second the main verb  then the
subcat list of the main verb instantiates the subcat list of the head,
and generation  becomes a deterministic procedure in which complements
are generated in sequence.  However, if the generator generates second
some complement  other than the main verb, then the subcat list of the
head  contains  no  restricting  information  to  guide  deterministic
generation, and  generation  becomes a  generate-and-test procedure in
which complements  are generated at random, only to  be  eliminated by
further unifications.  Clearly then,  the order of  evaluation of  the
complements  in  a rule  can  profoundly  influence the efficiency  of
generation, and an  efficient  head-driven  generator  must  order the
evaluation of the complements in a rule accordingly.

\subsection{Off-line versus On-line}
Dynamic, on-line reordering  can solve  the ordering problem discussed
in the previous  subsection, but is rather unattractive:  interpreting
grammar  rules  at  run  time   creates  much  overhead,  and  locally
determining  the  optimal   evaluation   order  is  often  impossible.
Goal-freezing can also  overcome the ordering  problem, but is equally
unappealing: goal-freezing is  computationally expensive,  it  demands
the  procedural  annotation   of  an   otherwise  declarative  grammar
specification,  and  it presupposes  that  a  grammar writer possesses
substantial  computational processing expertise.  We chose  instead to
deal  with the  ordering problem  by  using  off-line  compilation  to
automatically  optimize  a  grammar  such  that it  can  be  used  for
generation,  without  additional   provision  for  dealing   with  the
evaluation order, by our Earley generator.
\section{Off-line Grammar Optimization}
\label{off-line}
Our off-line grammar optimization is based on  a generalization of the
dataflow analysis employed in the {\sc dia} to a dataflow analysis for
typed  feature  structure  grammars.  This dataflow analysis  takes as
input a  specification of the  paths  of the  start category that  are
considered fully instantiated.  In case of generation, this means that
the user  annotates the path specifying the logical  form,  i.e.,  the
path {\footnotesize  cont} (or some of its subpaths), as  {\it bound}.
We use  the type  hierarchy and  an  extension  of the unification and
generalization operations such that path annotations are preserved, to
determine the flow of (semantic) information {\it  between} the  rules
and the  lexical entries in  a grammar.  Structure sharing  determines
the dataflow {\it within} the rules of the grammar.

The  dataflow analysis is used to determine the relative efficiency of
a particular evaluation order  of the right-hand side categories  in a
phrase structure rule by  computing  the  {\it  maximal  degree  of
nondeterminacy}
introduced by the evaluation of each of these categories.  The maximal
degree of nondeterminacy introduced by a right-hand side category equals the
maximal number  of  rules  and/or  lexical  entries  with  which  this
category   unifies   given  its  binding  annotations.    The  optimal
evaluation order of the right-hand side categories is found by comparing the
maximal degree of nondeterminacy introduced  by the evaluation  of the
individual categories with the degree of nondeterminacy the grammar is
allowed  to introduce: if the  degree of nondeterminacy introduced  by
the evaluation  of one  of  the right-hand side categories in a rule exceeds
the  admissible  degree  of  nondeterminacy the  ordering  at hand  is
rejected.  The  degree  of  nondeterminacy the  grammar is allowed  to
introduce is originally set to one and consecutively incremented until
the optimal evaluation order for all rules in the grammar is found.
\subsection{Example}
The compilation  process  is  illustrated on  the  basis  of the
phrase structure rule for argument composition discussed in
\ref{limitations}. Space limitations force  us  to abstract  over  the
recursive optimization of the rules defining  the right-hand side categories
through considering only the defining lexical entries.

Unifying the user  annotated start category with the left-hand side of
this phrase structure rule leads to the annotation of the path specifying the
logical form of the construction as bound (see below).  As a result of
the structure-sharing  between  the left-hand side of the rule and the
auxiliary  verb   category,  the  {\footnotesize  cont}-value  of  the
auxiliary  verb can be  treated  as bound, as well.   In addition, the
paths  with a value of a maximal  specific type for which there are no
appropriate features specified, for example,  the  path {\footnotesize
cat}, can be considered bound:

\begin{scriptsize}
\begin{center}
\begin{avm}
\[ cat$_{bound}$ & v\\
   subcat$_{bound}$ & \<\>\\
   cont$_{bound}$ & \@5
\]
$\longrightarrow$\end{avm}
\newline\begin{avm}
\[ cat$_{bound}$ & v\\
   fin$_{bound}$ & $+$\\
   aux$_{bound}$ & $+$\\
   subcat & \<\@3\|\@4\>\\
   cont$_{bound}$ & \@5
\]
,
\@1
,
\@2
,
\@3\[cat$_{bound}$ & v\\
     fin$_{bound}$ & $-$\\
     aux$_{bound}$ & $-$\\
     subcat & \@4\<\@1\|\@2\>
\]
\end{avm}
\end{center}
\end{scriptsize}
On  the  basis  of this annotated  rule,  we  investigate  the lexical
entries defining its  right-hand side categories.  The auxiliary verb
category  is  unified  with   its  defining   lexical  entries  (under
preservation of the binding annotations).  The following is an example
of  such  a lexical entry.  (Note that  subpaths  of  a path marked as
bound are considered bound too.)

\begin{scriptsize}
\begin{center}
\begin{avm}
\[ phon & hat\\
   cat$_{bound}$ & v\\
   fin$_{bound}$ & $+$\\
   aux$_{bound}$ & $+$\\
   subcat & \<\[cont$_{bound}$ & \@5\]\>\\
   \avmspan{cont$_{bound}$\|nucleus$_{bound}$\|arg$_{bound}$ & \@5}
\]
\end{avm}
\end{center}
\end{scriptsize}
The binding annotations of the lexical entries defining the  auxiliary
verb  are  used to determine with how many  lexical  entries  the right-hand
side category of the rule maximally unifies, i.e., its  maximal degree
of nondeterminacy. In this  case, the maximal degree of nondeterminacy
that the evaluation of the  auxiliary verb  introduces is very low  as
the   logical  form  of  the   auxiliary  verb  is   considered  fully
instantiated.   Now we mark the  paths of the defining lexical entries
whose instantiation can be deduced from the type hierarchy.   To mimic
the evaluation  of the  auxiliary  verb, we  determine the information
common to all defining lexical entries by taking their generalization,
i.e., the most specific feature structure subsuming all, and unify the
result  with the original  right-hand side category in  the  phrase
structure rule.
Because  both  the  generalization  and   the  unification  operations
preserve binding  annotations,  this  leads (via structure-sharing) to
the annotation  that the logical form of  the verbal complement can be
considered instantiated.  Note that  the nonverbal complements do  not
become  further  instantiated.   By subsequent  investigation  of  the
maximal degree of nondeterminacy introduced  by the evaluation  of the
complements in various  permutations, we find that the logical form of
a sentence  only restricts the evaluation of the nonverbal complements
after the evaluation of the  verbal  complement.  This can be verified
on the basis of a sample lexical entry for a main verb.

\begin{scriptsize}
\begin{center}
\begin{avm}
\[ phon &lieben\\
   cat & v\\
   fin & $-$\\
   aux & $-$\\
   subcat & \<\[cont \@6\;\],\[cont \@7\;\]\>\\
   \avmspan{cont\|nucleus\[lover \@6\;\\
                  loved \@7\;\]}
\]
\end{avm}
\end{center}
\end{scriptsize}
The  relative  efficiency of  this  evaluation leads our  compiler  to
choose
\begin{scriptsize}
\begin{center}
\begin{avm}
\[ cat & v\\
   subcat & \<\>\\
   cont & \@5
\]
$\longrightarrow$\end{avm}
\newline\begin{avm}

\[ cat & v\\
   fin & $+$\\
   aux & $+$\\
   subcat & \<\@3\|\@4\>\\
   cont & \@5
\]
,
\@3\[cat & v\\
     fin & $-$\\
     aux & $-$\\
     subcat & \@4\<\@1\|\@2\>
\]
,
\@1
,
\@2
\end{avm}
\end{center}
\end{scriptsize}
as  the  optimal evaluation order of  our phrase structure rule  for
argument  composition.

\subsection{Processing Head}
The optimal  evaluation  order for  a  phrase structure rule need  not
necessarily  be head-first. Our dataflow  analysis  treats  heads  and
complements  alike,  and  includes the head in  the calculation of the
optimal evaluation order of a rule.   If the evaluation of the head of
a  rule  introduces  much   nondeterminacy  or  provides  insufficient
restricting information  for the evaluation of  its  complements,  our
dataflow analysis might not select the  head as the first  category to
be evaluated, and choose instead

\begin{scriptsize}
\begin{center}
\begin{avm}
\[ cat & v\\
   subcat & \<\>\\
   cont & \@5
\]
$\longrightarrow$\end{avm}
\newline\begin{avm}

\@3\[cat & v\\
     fin & $-$\\
     aux & $-$\\
     subcat & \@4\<\@1\|\@2\>
\]
,
\[ cat & v\\
   fin & $+$\\
   aux & $+$\\
   subcat & \<\@3\|\@4\>\\
   cont & \@5
\]
,
\@1
,
\@2
\end{avm}
\end{center}
\end{scriptsize}
as  the  optimal  evaluation  order.   This  clearly  demonstrates  an
extremely  important  consequence of  using  our dataflow  analysis to
compile a declarative grammar into a grammar optimized for generation.
Empty  or displaced  heads  pose  us no  problem,  since  the  optimal
evaluation  order  of  the  right-hand side  of  a rule is  determined
regardless of the head.  Our dataflow analysis ignores the grammatical
head,  but  identifies  instead the `processing  head', and  (no  less
importantly) the `first processing complement', the `second processing
complement', and so on.
\section{Constraints on Grammar}
\label{pvp}

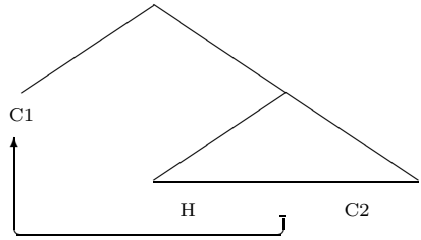
\begin{figure*}[htp]
\begin{scriptsize}
\begin{center}
\begin{picture}(253.5533,110)(0,0)
\put(95,97){\line(3,-2){50}}
\put(95,97){\line(-3,-2){50}}
\put(145,64){\line(3,-2){50}}
\put(145,64){\line(-3,-2){50}}
\put(195,30){\line(-1,0){100}}
\put(167,17){C2}
\put(142,17){\_}
\put(105,17){H}
\put(40,53){C1}
\put(42,12){\vector(0,0){35}}
\put(43,12){\oval(2,2)[bl]}
\put(43,10){\line(1,0){100}}
\put(143,12){\oval(2,2)[br]}
\put(144,12){\line(0,0){4}}
\end{picture}
\end{center}
\end{scriptsize}
\caption{Complement displacement.}
\label{fig1}
\end{figure*}

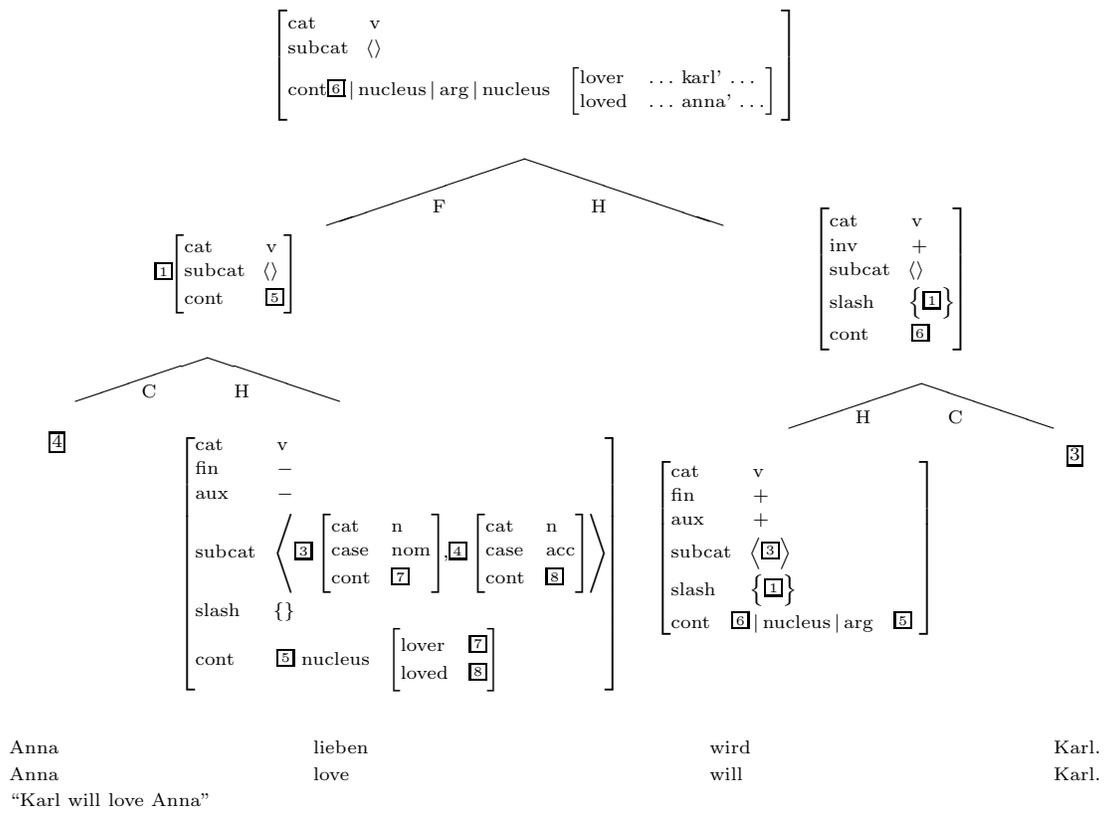
\begin{figure*}[htp]
\begin{scriptsize}
\begin{center}
\begin{picture}(453.5533,285)(0,0)
\put(120,285)
{\begin{avm}\[cat & v\\
              subcat &\< \>\\
    \avmspan{cont\@6\|nucleus\|arg\|nucleus\;\[ lover & $\ldots$ karl'
$\ldots$\\
                                          loved & $\ldots$ anna'
                                          $\ldots$   \]}
\]
\end{avm}}
\put(215,235){\line(3,-1){75}}\put(180,215){F}
\put(240,215){H}
\put(215,235){\line(-3,-1){75}}
\put(75,200)
{\begin{avm}
\@1\[cat & v\\
     subcat & \< \>\\
     cont & \@5
\]
\end{avm}}
\put(70,145){C}
\put(105,145){H}
\put(95,160){\line(3,-1){50}}
\put(95,160){\line(-3,-1){50}}
\put(325,210)
{\begin{avm}
\[ cat & v\\
   inv & $+$\\
   subcat & \< \>\\
   slash & \{ \@1 \}\\
   cont & \@6
\]
\end{avm}}
\put(340,135){H}
\put(375,135){C}
\put(365,150){\line(3,-1){50}}
\put(365,150){\line(-3,-1){50}}
\put(35,125)
{\begin{avm}
\reentry{4}
\end{avm}
}
\put(20,10){Anna}
\put(135,10){lieben}
\put(285,10){wird}
\put(415,10){Karl.}
\put(20,0){Anna}
\put(135,0){love}
\put(285,0){will}
\put(415,0){Karl.}
\put(20,-10){``Karl will love Anna''}
\put(85,125)
{\begin{avm}
\[ cat & v\\
   fin & $-$\\
   aux & $-$\\
   subcat & \<\@3
\[
cat & n\\
case & nom\\
cont & \@7\\
\],\@4
\[
cat &  n\\
case & acc\\
cont & \@8\\
\]\>\\
   slash & \{\}\\
   cont & \@5 nucleus\;\[ lover & \@7\\
                          loved & \@8   \]
\]
\end{avm}}
\put(265,115)
{\begin{avm}
\[ cat & v\\
   fin & $+$\\
   aux & $+$\\
   subcat & \<\@3\>\\
   slash & \{\@1\}\\
   \avmspan{cont\;\@6\|nucleus\|arg\;\@5} \]
\end{avm}}
\put(420,120)
{\begin{avm}
\reentry{3}
\end{avm}}
\end{picture}
\end{center}
\end{scriptsize}
\caption{Example of problematic complement displacement.}
\label{fig2}
\end{figure*}
Our Earley  generator and the described compiler  for off-line grammar
optimization  have  been extensively tested with  a  large  {\sc hpsg}
grammar.   This  test-grammar  is  based on  the  implementation of an
analysis of partial  {\sc  vp}  topicalization in German \cite{HMN} in
the   Troll  system  \cite{Gerdemann+King}.   Testing  the   developed
techniques uncovered  important constraints  on the form  of the
phrase structure rules in a grammar imposed by the compiler.

\subsection{Complement Displacement}
The  compiler is not  able to find an  evaluation order  such that the
Earley  generator has sufficient restricting information  to  generate
all subparts  of  the construction efficiently  in particular cases of
complement displacement.  More  specifically, this problem arises when
a complement receives  essential restricting information from the head
of the construction  from which it has been  extracted, while, at  the
same  time, it  provides  essential restricting  information  for  the
complements  that   stayed   behind.   Such  a  case   is  represented
schematically in  figure \ref{fig1} (see next page).

The first processing complement ({\sc c{1}}) of the head ({\sc h}) has
been  displaced.  This  is  problematic  in  case  {\sc c{1}} provides
essential  bindings for  the successful evaluation  of  the complement
{\sc c{2}}.  {\sc c{1}}  can not be  evaluated prior to  the  head and
once {\sc h} is evaluated  it  is  no longer possible to evaluate {\sc
c{1}}   prior   to    {\sc   c{2}}.    An   example   of   problematic
complement displacement taken from our test-grammar is given in figure
\ref{fig2} (see next page).   The topicalized partial {\sc  vp} {\footnotesize
``Anna
lieben''}  receives  its  restricting  semantic  information from  the
auxiliary verb and upon its evaluation provides essential bindings not
only for  the  direct  object,  but also for the  subject that  stayed
behind  in the Mittelfeld  together  with  the auxiliary verb.   These
mutual dependencies between the subconstituents of two different local
trees  lead either to the unrestricted  generation of the partial {\sc
vp},  or  to  the  unrestricted  generation  of  the  subject  in  the
Mittelfeld.  We handled this problem by partial execution
\cite{Pereira+Shieber}  of the  filler-head  rule.   This  allows  the
evaluation of  the filler  right after the evaluation of the auxiliary
verb, but prior to the subject.  A  head-driven  generator has to rely
on a similar solution, as it  will  not be able to  find a  successful
ordering for the local trees either, simply because it does not exist.

\subsection{Generalization}
A potential problem for our approach constitutes  the requirement that
the phrase structure rules in the grammar need to have a  particular degree of
specificity for  the generalization operation to be used  successfully
to mimic its evaluation. This is  best illustrated on the basis of the
following, more `schematic', phrase structure rule:

\begin{scriptsize}
\begin{center}
\begin{avm}
\[ cat & v\\
   subcat & \<\>\\
   cont & \@4
\]
$\longrightarrow$
\[ cat & v\\
   fin & $+$\\
   subcat & \<\@3,\@1,\@2\>\\
   cont & \@4
\]
,
\@1
,
\@2
,
\@3
\end{avm}
\end{center}
\end{scriptsize}
Underspecification of the  head of the  rule allows it  to  unify with
both  finite  auxiliaries  and  finite  ditransitive  main verbs.   In
combination  with  the  underspecification  of  the complements,  this
allows  the  rule  not  only  to  be  used  for  argument  composition
constructions, as discussed above, but also for constructions in which
a finite main verb becomes saturated. This means that the logical form
of  the  nonverbal  complements  ({\footnotesize  $\reentry{1}$}   and
{\footnotesize  $\reentry{2}$})  becomes  available  either  upon  the
evaluation of  the complement tagged {\footnotesize $\reentry{3}$} (in
case  of argument composition), or upon the  evaluation of  the finite
verb (in case the head of the rule is a ditransitive main verb).  As a
result,  the  use  of generalization  does  not  suffice to mimic  the
evaluation of the respective right-hand side categories.  Because both
verbal  categories  have  defining   lexical  entries  which   do  not
instantiate the logical form  of the nonverbal arguments, the dataflow
analysis  leads  to  the  conclusion  that  the  logical  form of  the
nonverbal complements  never becomes instantiated.   This  causes  the
rejection of  all  possible evaluation orders  for this  rule,  as the
evaluation of an unrestricted nonverbal complement clearly exceeds the
allowed  maximal  degree  of nondeterminacy of  the  grammar.  We  are
therefore  forced  to split this  schematic phrase structure rule into
two more specific rules at least  during the optimization process.  It
is  important  to  note  that this  is  a  consequence  of  a  general
limitation of dataflow analysis (see also Mellish, 1981).
\section{Concluding Remarks}
An innovative approach to {\sc hpsg} processing is described that uses
an off-line compiler to automatically prime  a declarative grammar for
generation  or parsing,  and inputs the  primed grammar to an advanced
Earley  processor.   Our  off-line  compiler  extends  the  techniques
developed in the context of the {\sc  dia}  in that it  compiles typed
feature structure  grammars, rather than simple  logic grammars.   The
approach  allows  efficient  bidirectional   processing  with  similar
generation  and  parsing times. It is shown  that  combining  off-line
techniques with an advanced Earley-style generator provides an elegant
solution to the general problem that empty or displaced heads pose for
conventional head-driven generation.

The developed off-line compilation  techniques make crucial use of the
fundamental properties of the {\sc hpsg} formalism.   The monostratal,
uniform treatment of syntax, semantics and phonology supports dataflow
analysis, which is used extensively to provide  the  information  upon
which  off-line compilation  is  based.  Our  compiler  uses  the type
hierarchy to determine paths with a  value of  a minimal type  without
appropriate features  as bound.   However, the equivalent of this kind
of minimal types in  untyped feature structure grammars  are constants
which can be used in a similar fashion for off-line optimization.

\end{document}